\documentclass[10pt,twocolumn,letterpaper]{article}

\usepackage[pagenumbers]{cvpr} 

\usepackage{times}
\usepackage{epsfig}
\usepackage{graphicx}
\usepackage{amsmath}
\usepackage{amssymb}

\usepackage{booktabs}
\usepackage{balance}
\usepackage{multirow}
\usepackage{multicol}
\usepackage{colortbl}
\definecolor{mygray}{gray}{.85}
\definecolor{myhighlight}{RGB}{193,210,240}
\usepackage{mwe}
\usepackage{algorithmic,float}
\usepackage[linesnumbered,ruled,vlined,noline]{algorithm2e}
\usepackage{etoolbox}
\usepackage{relsize}
\usepackage{comment}
\usepackage{bm}
\usepackage{enumitem}
\usepackage{adjustbox}
\usepackage{soul}
\usepackage[dvipsnames]{xcolor}

\definecolor{cvprblue}{rgb}{0.21,0.49,0.74}
\usepackage[pagebackref,breaklinks,colorlinks,citecolor=cvprblue]{hyperref}

\def\paperID{8991} 
\def\confName{CVPR}
\def\confYear{2024}

\begin{document}

\title{Memory-based Adapters for Online 3D Scene Perception}

\author{Xiuwei Xu\textsuperscript{1}\thanks{}, ~Chong Xia\textsuperscript{1}\footnotemark[1], ~Ziwei Wang\textsuperscript{2}, ~Linqing Zhao\textsuperscript{3}, ~Yueqi Duan\textsuperscript{1}, ~Jie Zhou\textsuperscript{1}, ~Jiwen Lu\textsuperscript{1}\thanks{}\\
\textsuperscript{1}Tsinghua University, ~\textsuperscript{2}Carnegie Mellon University, ~\textsuperscript{3}Tianjin University \\
}

\twocolumn[{
\maketitle
\vspace{-8mm}
\begin{figure}[H]
\hsize=\textwidth
\centering
\includegraphics[width=2.1\linewidth]{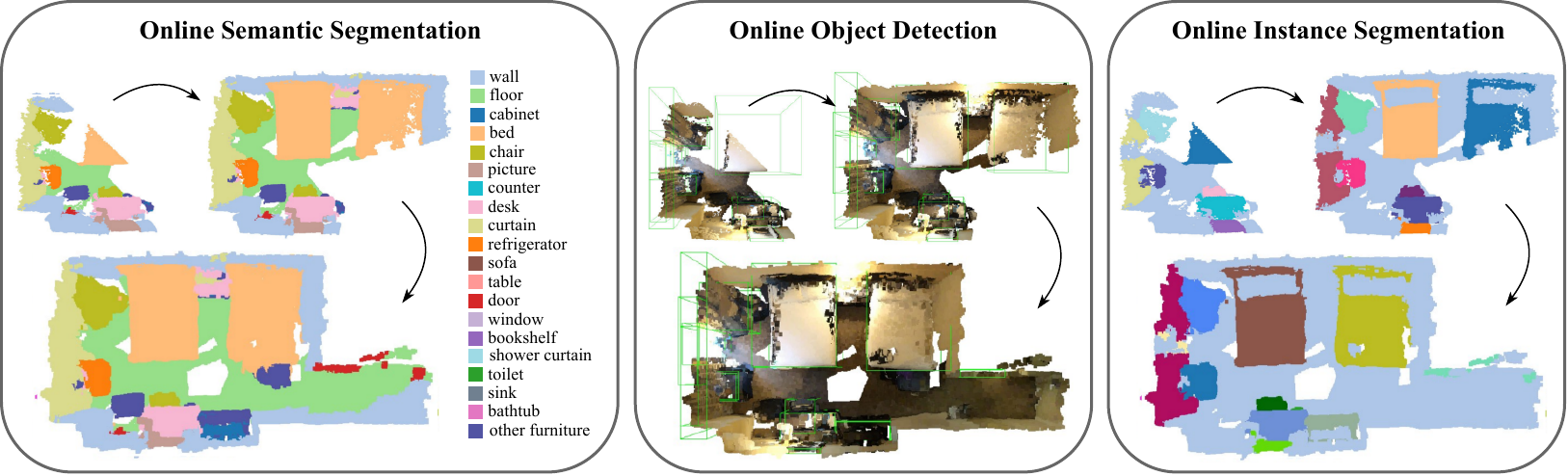}
\caption{We propose a general framework for online 3D scene perception. With the presented memory-based adapters, we empower existing offline models in different tasks with online perception ability, which is valuable for robotics applications.}
\label{teaser}
\end{figure}
}]
{
  \renewcommand{\thefootnote}%
    {\fnsymbol{footnote}}
  \footnotetext[1]{Equal contribution. ~~\textsuperscript{\dag}Corresponding author.}
}

\begin{abstract}
  In this paper, we propose a new framework for online 3D scene perception. 
Conventional 3D scene perception methods are offline, i.e., take an already reconstructed 3D scene geometry as input, which is not applicable in robotic applications where the input data is streaming RGB-D videos rather than a complete 3D scene reconstructed from pre-collected RGB-D videos.
To deal with online 3D scene perception tasks where data collection and perception should be performed simultaneously, the model should be able to process 3D scenes frame by frame and make use of the temporal information.
To this end, we propose an adapter-based plug-and-play module for the backbone of 3D scene perception model, which constructs memory to cache and aggregate the extracted RGB-D features to empower offline models with temporal learning ability. Specifically, we propose a queued memory mechanism to cache the supporting point cloud and image features. Then we devise aggregation modules which directly perform on the memory and pass temporal information to current frame. We further propose 3D-to-2D adapter to enhance image features with strong global context. 
Our adapters can be easily inserted into mainstream offline architectures of different tasks and significantly boost their performance on online tasks. 
Extensive experiments on ScanNet and SceneNN datasets demonstrate our approach achieves leading performance on three 3D scene perception tasks compared with state-of-the-art online methods by simply finetuning existing offline models, without any model and task-specific designs.
\href{https://xuxw98.github.io/Online3D/}{Project page}.
\end{abstract}

\section{Introduction}

3D scene perception aims to parse a 3D scene into semantic or object-level entities, mainly including semantic segmentation, object detection and instance segmentation, which is the fundamental ability for robotics or AR/VR applications. 
Since PointNet~\cite{qi2017pointnet} proposes the first model that directly process point clouds, great improvement on 3D scene perception~\cite{choy20194d,rukhovich2022fcaf3d,wang2022cagroup3d,schult2022mask3d,vu2022softgroup} has been achieved in recent years by accurate and efficient architecture design.

However, conventional 3D scene perception methods are offline, i.e., they take an already reconstructed 3D scene geometry from pre-collected RGB-D videos without temporal information as input. While in most robotic application such as navigation~\cite{chaplot2020object,zhang20233d} and manipulation~\cite{Mousavian_2019_ICCV} where the agent is usually initialized in an unknown environment, the input data is streaming RGB-D videos and scene perception should be performed synchronously with data collection to guide the agent how to explore. Therefore, online 3D scene perception model with temporal modeling ability is required, which takes in streaming RGB-D video and continuously outputs the perception of the currently observed 3D scene.
There are also a few online 3D scene perception methods~\cite{mccormac2017semanticfusion,narita2019panopticfusion,zhang2020fusion,huang2021supervoxel,liu2022ins} designed for special architecture and task. 
As these methods only focus on temporal aggregation for single modality, they cannot fully exploit temporal relations between image and point cloud features and thus their performances are not satisfactory. 

In this paper, we propose a new general framework for online 3D scene perception. Different from previous works which design online perception approachs based on specific architecture and task and train models on RGB-D videos from scratch, we convert exsiting offline 3D perception models to online ones by simply inserting plug-and-play modules and finetuning.
Taking inspiration from adapters~\cite{pan2022st,chen2022vision} which adapt image backbones to downstream tasks by additional parameter tuning, we propose memory-based adapters to empower the backbone of 3D perception model with temporal modeling ability by reusing the extracted features from previous frames. 
Specifically, we propose a queued memory mechanism to cache the supporting point cloud and image features for the RGB-D frame at current time. Based on the structure of memory, we devise aggregation modules which directly operate on the memory and pass temporal information to current frame. As the global context of image features is limited, we further propose 3D-to-2D adapter to enhance image features with 3D memory projection and 2D sparse aggregation.
In this way, we can make use of the existing mainstream 3D scene perception model zoo to acquire a series of online models, with simple inserting and finetuning.
We conduct extensive experiments on three online perception tasks on ScanNet~\cite{dai2017scannet} and SceneNN~\cite{hua2016scenenn} datasets.
Our approach achieves leading performance on all tasks and datasets without any additional loss function and special prediction fusion strategy.
To summarize, our contributions include:
\begin{itemize}
    \item We propose a new framework for online 3D scene perception, which extends existing offline models to online ones by adapter without model and task-specific design.
    \item We propose general memory-based adapters for image and point cloud backbones, which cache and aggregate extracted features to model the temporal relations between frames.
    \item Equipped with our adapters, offline models are able to achieve leading performance on three tasks compared with state-of-the-art online models.
\end{itemize}
  
\section{Related Work}
  \textbf{3D Scene Perception:}
3D scene perception is widely studied in computer vision, which can be divided into three mainstream tasks: semantic segmentation~\cite{qi2017pointnet,qi2017pointnet++,graham20183d,choy20194d}, object detection~\cite{hou20193d,Qi_2019_ICCV,rukhovich2022fcaf3d,wang2022cagroup3d} and instance segmentation~\cite{yi2019gspn,jiang2020pointgroup,vu2022softgroup,schult2022mask3d,kolodiazhnyi2023top}.
As we focus on the feature extraction of 3D scene in this work, we mainly discuss the backbone of 3D scene perception networks. Due to the unordered property of point cloud data, voxelizing the points and applying convolution on 3D grids is a natural solution~\cite{chang2015shapenet,qi2016volumetric}.
However, the computational cost and memory requirement both increase cubically with voxel resolution, which is inefficient.
PointNet~\cite{qi2017pointnet} is the pioneer work which directly extracts feature representations from raw point clouds. PointNet++ proposes set abstraction and feature propagation operation based on PointNet, which helps learning more detailed local geometric information. As the furthest point sampling operation in PointNet++ is time consuming, PV-CNN~\cite{liu2019point} converts point clouds to low-resolution voxels and apply 3D convolution to efficiently aggregate local features. Another way to extract high-quality 3D features is sparse convolution~\cite{graham2014spatially,engelcke2017vote3deep}, which voxelizes the point clouds but only apply 3D convolution on non-empty voxels. To further improve the efficiency of sparse CNNs, submanifold sparse convolution~\cite{graham20183d,choy20194d} is introduced, which only conducts convolution when the center of kernel slides over active sites and keeps the same level of sparsity throughout the network. However, these methods are designed for offline 3D scene perception, which is not able to process a streaming RGB-D video at real time.

\begin{figure*}[t]
    \centering
    \includegraphics[width=0.9\linewidth]{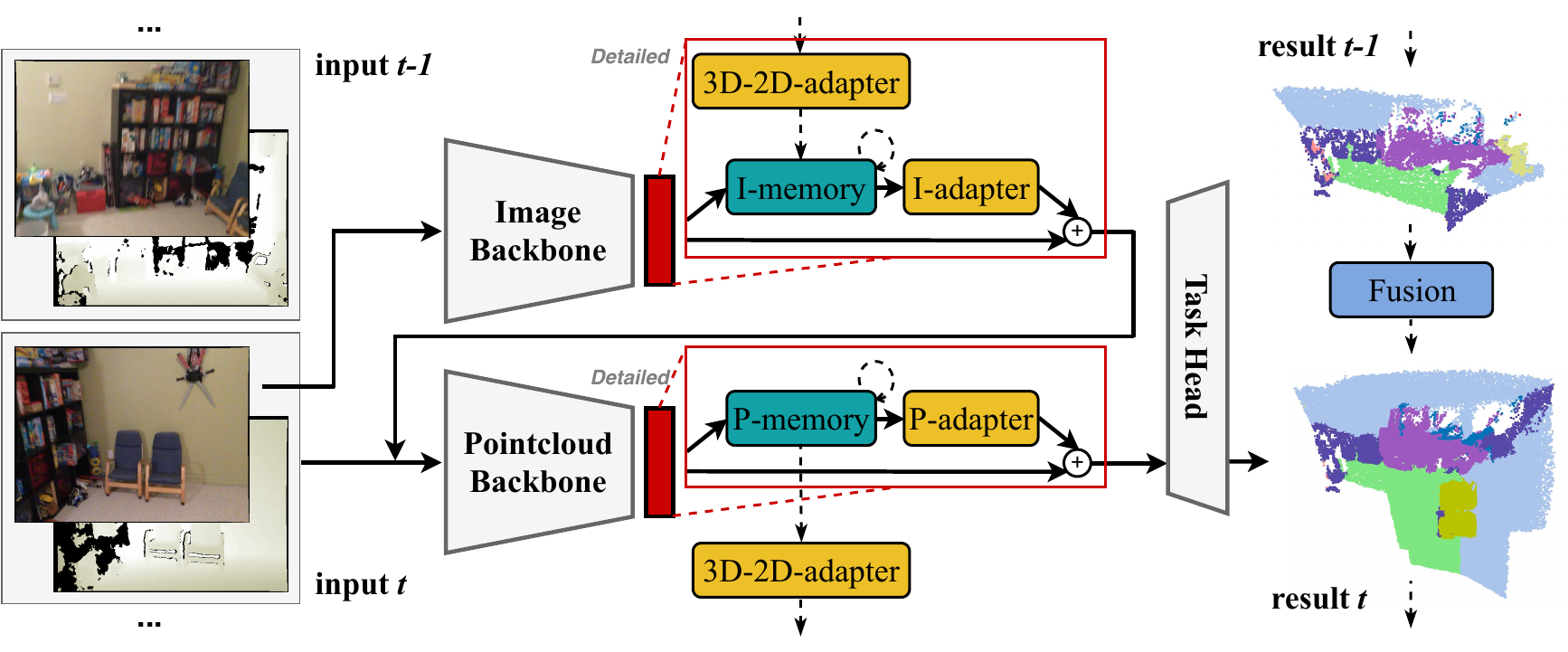}
    \caption{Overall architecture of our approach. We insert memory-based adapters after image and point cloud backbones, which cache the extracted features in memory over time and perform temporal aggregation. 3D-to-2D adapter is proposed to further exploit inter-modal temporal information. 
    Solid lines indicate operations within a single frame, while dashed lines indicate temporal operations.}
    \label{over-arch}
\end{figure*}

\textbf{Streaming Data Analysis:}
As the input for online 3D scene perception model is streaming RGB-D video, we review the streaming data analysis methods for both image and point cloud domains.
In 2D vision, many works~\cite{carreira2018massively,dai2019transformer,kondratyuk2021movinets} extend causal convolution~\cite{oord2016wavenet} to streaming videos, where a stream buffer is devised to cache previous frames and 3D causal convolution is applied to unidirectionally aggregate spatial-temporal information. 
TSM~\cite{lin2019tsm} utilizes a more efficient shift mechanism, where a proportion of channels of previous image features are shifted to the next frame. Then the spatial-temporal information can be efficiently aggregated by 2D convolution. Our image adapter also utilizes channel shift for efficient temporal modeling. Differently, TSM is trained from scratch where the network can learn how to model temporal information according to the shift proportion. While we reorganize the channels and adopt channel shift in a plug-and-play adapter to empower image backbone with temporal modeling ability.
However, as 2D streaming videos contain less helpful information for real world applications like robotic navigation~\cite{chaplot2020object,zhang20233d} and manipulation~\cite{Mousavian_2019_ICCV}, increasing attention has been paid to 3D streaming RGB-D video analysis.
A natural solution is to first process 2D images and then project the predictions to 3D point clouds, which is followed by a fusion step to merge the predictions from different frames~\cite{mccormac2017semanticfusion,narita2019panopticfusion}.
Fusion-aware 3D-Conv~\cite{zhang2020fusion} and SVCNN~\cite{huang2021supervoxel} maintain the information of previous frames in 3D space and conduct point-based convolution to fuse the 3D features for semantic segmentation. INS-CONV~\cite{liu2022ins} extends sparse convolution to incremental CNN to efficiently extract global 3D features for semantic and instance segmentation.
Differently, our approach empowers offline model with online perception ability by image and point cloud memory-based adapters, which fully exploits the multimodal temporal relations.

\section{Approach}

In this section, we first introduce the definition of online 3D scene perception and explain our motivation of memory-based adapter. Then we describe how to construct memory and refine the backbone features by adapter for point cloud and image respectively.

\subsection{Online 3D Scene Perception}\label{def}
Let $\mathcal{X}_t=\{x_1,x_2,...,x_t\}$ be a posed RGB-D streaming video, which means the video is collected with the moving of sensor rather than a pre-collected video. We have:
\begin{equation}
    x_t=(I_t,P_t,M_t),\ I_t\in \mathbb{R}^{H\times W\times 3},\ P_t\in \mathbb{R}^{N\times 3},\ M_t\in \mathbb{R}^{3\times 4}
\end{equation}
where $I_t$ and $P_t$ refer to the image and point clouds for one RGB-D frame. $P_t$ is acquired by lifting the depth image to world coordinate system with pose parameters $M_t$, where $M_t$ can be estimated by visual odometry~\cite{park2017colored,zhao2021surprising}.
At time $t$, the input to online perception model is $\mathcal{X}_t$ and the output is predictions for the observed 3D scene $S_t=\bigcup_{i=1}^t{P_i}$, which can be bounding boxes and semantic/instance masks. Some works also perform 3D reconstruction along with online perception~\cite{liu2022ins} to acquire high-quality point clouds or meshes, but this is not required~\cite{chaplot2020object,ramakrishnan2022poni,zhang20233d}. In this work we do not rely on 3D reconstruction and directly take RGB-D streaming video as input, which is a more general setting.

Although great improvement has been achieved in the design of 3D perception models, most of them only focus on two scenarios: (1) Reconstructed scenes~\cite{armeni20163d,dai2017scannet}. The model $\mathcal{M}_{Rec}$ is trained on point clouds of complete scenes reconstructed from RGB-D videos. (2) Single-view scenes~\cite{Silberman2012nyu,song2015sun}. The model $\mathcal{M}_{SV}$ is trained on single-view point clouds back-projected from RGB-D image. However, $\mathcal{M}_{Rec}$ requires the input to be a complete scene, which is not accessible in real-time tasks. $\mathcal{M}_{SV}$ is able to process RGB-D videos frame-by-frame, but fails to exploit temporal information. 
Therefore, previous 3D models are not ready for the more practical online scene perception.

To this end, we aim to devise a plug-and-play temporal learning module, which can be inserted into any single-view perception model $\mathcal{M}_{SV}$ and empowers it with temporal modeling ability. 
Note that $\mathcal{M}_{SV}$ is originally a 3D perception model. We can extend it to a RGB-D perception model by early fusing the image features to point clouds~\cite{rukhovich2023tr3d}:
\begin{gather}
    p_t=\mathcal{M}_{SV}(P'_t), \nonumber \\
    P'_t=P_t\oplus \mathcal{S}(\mathcal{M}_I(I_t),P_t,M_t),\ \mathcal{S}(\cdot)\in \mathbb{R}^{N\times C}
\end{gather}
where $p_t$ is the prediction for input frame at time $t$. $\mathcal{M}_I$ is a image backbone pretrained on the same perception task as $\mathcal{M}_{SV}$. $\mathcal{S}$ projects $P_t$ to image coordinate system by $M_t$ and samples the corresponding 2D features to enrich the features of $P_t$.
We divide $\mathcal{M}_{SV}$ into a backbone $\mathcal{M}_{P}$ for extracting features of point clouds and a task-specific head $\mathcal{M}_{H}$. The goal of this work is to construct an image memory-based adapter for $\mathcal{M}_I(I_t)$ and a point cloud memory-based adapter for $\mathcal{M}_{P}(P'_t)$ to store and reuse the extracted backbone features for temporal modeling:
\begin{gather}
    \mathcal{M}_{I}(I_t),m^I_t=\mathcal{A}_I(\mathcal{M}_{I}(I_t),m^I_{t-1},m^P_{t-1}), \nonumber \\
    \mathcal{M}_{P}(P'_t),m^P_t=\mathcal{A}_P(\mathcal{M}_{P}(P'_t),m^P_{t-1})
\end{gather}
where the memory-based adapter $\mathcal{A}$ updates the memory $m_t$ with current features and refines current features by reusing the memory. 
We fully exploit inter-modal and intra-modal relationships: $\mathcal{M}_{SV}$ fuses image features to point clouds, $\mathcal{A}_P$ reuses previous point cloud features to refine current point cloud features, and $\mathcal{A}_I$ reuses previous features from both modal to refine the current image features.
As shown in Figure \ref{over-arch}, we follow the same paradigm to design both modules, which can be easily embedded into image and point cloud backbones to achieve temporal modeling.

\subsection{Temporal Modeling for Point Clouds}
Given $\{\mathcal{M}_{P}(P'_{1}),\mathcal{M}_{P}(P'_{2}),...,\mathcal{M}_{P}(P'_{t})\}$ at time $t$, i.e., the sequence of point cloud features extracted from the backbone, we aim to enhance the current features $\mathcal{M}_{P}(P'_{t})$ by exploiting the temporal relations within this sequence. Here we first construct a memory to efficiently cache the point cloud features from different timestamp. Then we aggregate the temporal information from the memory to features at current time $t$ with a plug-and-play adapter.

\begin{figure}[t]
    \centering
    \includegraphics[width=1.0\linewidth]{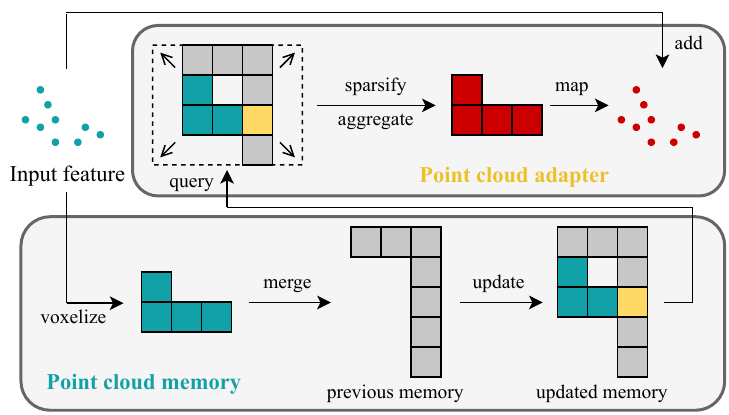}
    \caption{The architecture of the memory-based adapter for point cloud features. We cache and aggregate the features in a queue of 3D voxel grids. Gray, green, yellow and red block refer to previous, current, updated and aggregated voxel features.}
    \label{point-module}
\end{figure}

\textbf{Memory construction:}
The temporal information for 3D scenes is reflected in a more complete geometry. As a single RGB-D frame may not contain a complete large objects or high-level scene context, the geometric information from previous frames are important for accurate perception of current frame. Therefore, we can cache the sequence of extracted point cloud features in a shared 3D space. A simple way is to directly store the features in terms of point clouds in the world coordinate system. However, this is inefficient in both storage and computation: (1) as the coordinates of the point cloud are real values, the storage overhead will keep growing even if the RGB-D camera is not moving; (2) the number of points will be very large over time, so point-based sampling and feature aggregate take up high computation overhead.
To this end, we propose to store the features in a quantitized coordinate system, where point clouds are voxelized and stored in 3D grids. We also maintain the voxels in a queue to reduce the memory footprint when the scene is too large.
Specifically, $\mathcal{M}_{P}(P'_t)$ is first voxelized into the voxel grids $V_t$ by averaging all features whose coordinates falls into the same grid. These voxels are tagged with timestamp $t$. Then we merge $V_t$ to the memory $m^P_{t-1}$ by maxpooling:
\begin{gather}
    m^P_{t}={\rm maxpooling}(V_t,m^P_{t-1}), \nonumber \\
    m^P_{t}={\rm deq}(m^P_{t}, l)\ \ if\ \ N(m^P_{t})>N_{max}
\end{gather}
where ${\rm maxpooling}$ refers to channel-wise maxpooling conducted on each voxel grid, which will update both features and timestamps. ${\rm deq}(\cdot,l)$ means removing voxels with timestamp earlier than $t-l+1$ from the memory. $N(m^P_{t})$ is the number of voxels in the memory.
We utilize maxpooling as it preserves the most discriminative features over time, which is also efficient to compute as only features at time $t-1$ and $t$ are required.

\textbf{Memory-based adapter:}
After caching and updating the point cloud features in voxels, we need to make use of $m^P_{t}$ to enhance $\mathcal{M}_{P}(P'_t)$ with temporal information.
To exploit rich scene context from the memory $m^P_t$ while reduce redundant computation, we first use the coordinates of $V_t$ to query a neighbor voxel set:
\begin{equation}
    \mathcal{N}(V_t)=\{m^P_t[x][y][z]|(x,y,z)\in s*\mathcal{B}(V_t)\}
\end{equation}
where $\mathcal{N}(V_t)$ is the queired neighborhood of $V_t$. $m^P_t[x][y][z]$ refers to the voxel in $m^P_t$ at coordinate $(x,y,z)$. $\mathcal{B}$ is the minimum enclosing axis-aligned bounding box of $V_t$ and $s$ is a scaling factor to enlarge the size of box. In this way, the temporal information which provides supporting geometric information for current frame is collected into this voxel set. 

We then convert $\mathcal{N}(V_t)$ to a sparse tensor~\cite{graham20183d,choy20194d}, which is followed by a 3D sparse convolutional module $\mathcal{A}_P$ to aggregate the context information within $\mathcal{N}(V_t)$ to locations of $V_t$.
Finally we update $\mathcal{M}_{P}(P'_t)$ in an adapter-manner: (1) We map the aggregated features back to the coordinates of $\mathcal{M}_{P}(P'_t)$ and then add it to the original features with residual connection; (2) The adapter module $\mathcal{A}_P$ is zero-initialized. Therefore after inserting the adapter, finetuning will smoothly start from the original point.

\subsection{Temporal Modeling for Images}
For the sequence of image features $\{\mathcal{M}_{I}(I_1),...,\mathcal{M}_{I}(I_t)\}$ at time $t$, we follow the same paradigm as the point cloud counterpart to store it with a memory and aggregate temporal information to current frame $\mathcal{M}_{I}(I_t)$ with an adapter.

\begin{figure}[t]
    \centering
    \includegraphics[width=1.0\linewidth]{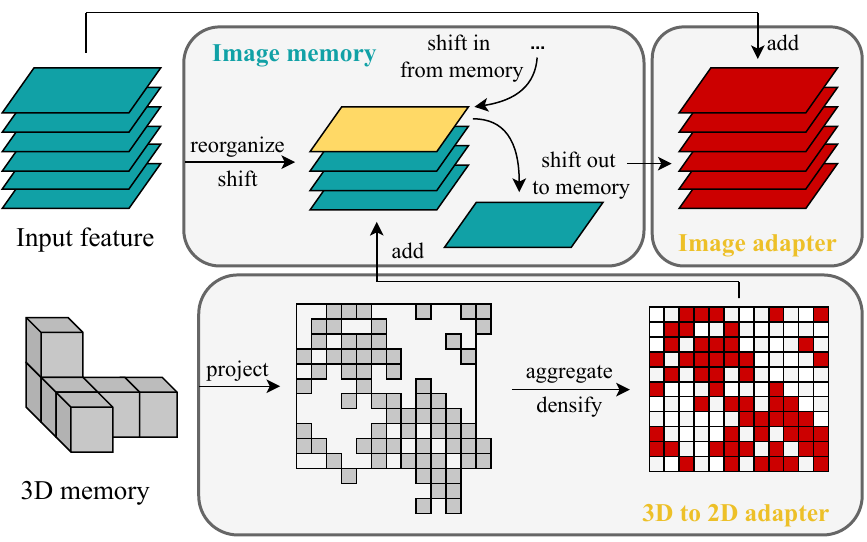}
    \caption{The architecture of the memory-based adapter for image features. We reorganize the input features and shift out a proportion of channels into the memory, while shifting in previous memory and aggregating temporal information by 2D convolution. We also resort to the 3D memory for more global context.}
    \label{img-module}
\end{figure}

\textbf{Memory construction:}
Different from 3D data where point clouds from different timestamps can be stored in a shared 3D space, for 2D data the image features are stacked into a streaming video.
A common practice to process this kind of data~\cite{kondratyuk2021movinets} is to maintain a queue and perform causal convolution (unidirectional along temporal dimension) to aggregate the information from previous frames to current image features. However, video analysis methods focus on extracting the global information of the streaming video up to now, while in our case we only need to enhance the current image features $\mathcal{M}_{I}(I_t)$. So causal convolution on previous frames will bring a large amount of redundant computation. 
Moreover, in online 3D scene perception, the most important information for image features is the observation of objects from multiple perspectives. Since an object is usually observed in a few adjacent frames, maintaining a short queue is enough in most cases.
To this end, we make an extreme simplification: we only store one frame of previous image and exploit temporal relations between frames at time $t-1$ and $t$.
In order to efficiently aggregate the adjacent frames, we adopt channel shift to cache the temporal information. Formally, given $\mathcal{M}_{I}(I_t)\in \mathbb{R}^{H\times W\times C}$, we learn a linear transformation $R_1\in \mathbb{R}^{C\times C'}$ to map the image features into another embedding space, where the first $\frac{1}{\tau}$ channels contain rich temporal information relevant to the next frame. Therefore the memory can be simply constructed by shifting out this part of channels:
\begin{gather}
    m^I_t=(\mathcal{M}_{I}(I_t)\cdot R_1)_{[:,:,:\frac{C'}{\tau}]}
\end{gather}
Note that this operation is repeated frame-by-frame, thus $m^I_{t-1}$ contains the temporal information relevant with current features $\mathcal{M}_{I}(I_t)$.

\textbf{Memory-based adapter:}
At time $t$, after shifting out a proportion of channels to memory, we can pad the empty channels with previous memory $m^I_{t-1}$. In this way, the temporal information of the adjacent two frames are merged into a single frame, for which we can directly adopt 2D convolution to aggregate the features:
\begin{gather}
    F_t={\rm 2D\text{-}Conv}(m^I_{t-1}\oplus(\mathcal{M}_{I}(I_t)\cdot R_1)_{[:,:,\frac{C'}{\tau}:]})\cdot R_2
\end{gather}
where $R_2\in \mathbb{R}^{C'\times C}$ is a learnable inverse transformation to map image features back to the original embedding space.
Finally we update $\mathcal{M}_{I}(I_t)$ by adding $F_t$ with a residual connection. We also zero-initialize $R_1$, $R_2$ and ${\rm 2D\text{-}Conv}$ for smooth finetuning.

\subsection{Inter-modal Temporal Modeling}
Although maintaining a short queue and adopting channel shift are able to effectively aggregate temporal information for image features, the global context is limited. This will lead to a performance drop when an object is very large or the RGB-D camera stops moving.
However, as analyzed before, extracting rich global context from a streaming video is really memory and computation consuming.
To solve this problem, we resort to the point cloud memory for global context extraction, because the point cloud features are efficiently cached in a shared 3D space and thus the length of queue can be long. We devise a 3D-to-2D adapter to refine current image features with the global 3D features.
Here we first project $m^P_{t-1}$ to the discrete image coordinate system with the inverse function of $\mathcal{S}$, which are then converted to sparse tensor. 
In this way, we keep the sparsity of point cloud memory and make the 2D features geometry-aware.
2D sparse convolution is then applied on the sparse tensor to aggregate the context information, followed by a densify operation to keep features inside image and pad other pixels with zero. 
Finally we add the densified 2D features to enhance $\mathcal{M}_{I}(I_t)\cdot R_1$ with richer global context. We zero-initialize the 2D sparse convolution for smooth training.

To acquire the final online perception results, a prediction fusion strategy is needed. As the focus of our work is the temporal learning modules for image and point cloud backbones, we only adopt a simple post-processing strategy to fuse the predictions of each frame in a whole, which we detail in Section \ref{impd}.





  
\section{Experiment}

In this section, we first describe our datasets and implementation details. Then we compare our method with state-of-the-art methods on both room-level and online benchmarks to comprehensively analyze the advantage of memory-based adapters. Finally we conduct ablation studies to validate the effectiveness of our design.

\subsection{Benchmarks and Implementation Details}\label{impd}
We evaluate our method on two datasets: ScanNet~\cite{dai2017scannet} and SceneNN~\cite{hua2016scenenn}. ScanNet contains 1513 scanned scene sequences, out of which we use 1201 sequences for training and the rest 312 for testing. SceneNN is a smaller dataset which contains 50 high-quality scanned scene sequences with semantic label. After careful filtering, we select 12 clean sequences for testing. We train all models on ScanNet and evaluate them on ScanNet or SceneNN.

\textbf{Benchmarks:} We first compare different methods on room-level benchmarks, i.e., the performance on the reconstructed complete scenes. 
For semantic segmentation, we compare different methods on ScanNet and SceneNN. Since online methods may not perform 3D reconstruction, we map their predictions on point clouds concatenated from posed RGB-D images to the reconstructed point clouds with nearest neighbor interpolation. For object detection and instance segmentation, the metric is computed on each object rather than the whole point clouds. Therefore we use reconstructed point clouds and RGB-D videos as the inputs for offline and online methods respectively, and calculate metrics based on their respective inputs.

We also follow AnyView~\cite{wu2023anyview} to organize an online benchmark on ScanNet for more comprehensive evaluation. We divide the RGB-D video of each room into several non-overlapping sequences and regard each sequence as an independent scene, where the number or the length of each sequence can be set to different values. In this way, we can measure the generalization ability of different methods when the input scenes are incomplete and of variable scales, which is a more practical setting. In our experiments, we divide each room into 1/5/10 sequences or sequences with fixed length 5/10/15, resulting in 6 metrics.

\textbf{Implementation details:} To train $\mathcal{M}_{SV}$, we first train a 2D perception model $\mathcal{M}_{I}$ following Pri3D~\cite{hou2021pri3d}. We use UNet~\cite{ronneberger2015u} for semantic segmentation and Faster-RCNN~\cite{ren2016faster} (only ResNet~\cite{he2016deep} and FPN~\cite{lin2017fpn} backbones are needed) for object detection and instance segmentation. Then we fix the image backbone and train $\mathcal{M}_{SV}$ on ScanNet-25k~\cite{dai2017scannet}, which is a single-view RGB-D dataset. For online perception, we zero-initialize the memory-based adapters and insert them into $\mathcal{M}_{SV}$. Then we train the new model on RGB-D videos from ScanNet. To reduce memory footprint, we randomly sample 8 adjacent RGB-D frames for each scene at every iteration.
We insert our memory-based adapters between the backbone and neck. For backbones which output multi-level features, we insert different adapters to different levels. In terms of hyperparameters, we set $l=50$, $s=2.5$, $\tau=8$ and $\delta=0.03$. We simply use the same optimizer configurations to train the models as in their original paper (designed for offline training).

For prediction fusion, we adopt different strategies for different tasks. 
\emph{Semantic segmentation}: The predictions for each frame are concatenated, which has the same point number with $S_t$. We use 2$cm$ voxelization to unify the predictions for points inside the same voxel grid by channel-wise maxpooling. 
\emph{Object detection}: The predicted bounding boxes for each frame are merged by 3D NMS. When two boxes of different time are colliding during NMS, we add $\delta$ to the classification scores of box in the newer frame. This is because our method ensures the backbone extracts more complete geometric features for the newer frame.
\emph{Instance segmentation}: Instance segmentation can be divided into transformer-based~\cite{schult2022mask3d}, grouping-based~\cite{jiang2020pointgroup,vu2022softgroup} and detection-based~\cite{hou20193d,yi2019gspn,kolodiazhnyi2023top}. As the first two kinds require a specially designed mask fusion strategy~\cite{liu2022ins} for online perception, we opt for the detection-based manner, where we can first conduct online 3D object detection and then utlize the boxes to crop and segment the point cloud features stored in the memory.

\subsection{Comparison with State-of-the-art}
We compare our method with the top-performance offline and online 3D perception models. Offline models refer to $\mathcal{M}_{Rec}$ described in Section \ref{def}, which is trained on reconstructed point clouds. Models with suffix "-SV" refer to $\mathcal{M}_{SV}$ that is trained on single-view RGB-D images.

\begin{table}[t]
    \setlength{\tabcolsep}{4pt}
    \centering
    \caption{3D semantic segmentation results on ScanNet and SceneNN datasets. For online methods, we map the predictions on point clouds concatenated from posed RGB-D images to the reconstructed point clouds to compare with offline method.}\label{tab1}
    \begin{tabular}{lc|cc|cc}  
        \toprule
        \multirow{2}{*}{Method} & \multirow{2}{*}{Type} & \multicolumn{2}{c|}{ScanNet} & \multicolumn{2}{c}{SceneNN} \\
        & & mIoU & mAcc & mIoU & mAcc \\
        \midrule
        MkNet~\cite{choy20194d} & Offline &71.6 &80.4 &-- &-- \\
        \midrule
        Fs-A~\cite{zhang2020fusion} & Online &63.5 &73.7 &51.1 &62.4 \\
        \rowcolor{mygray} MkNet-SV & Online &68.8 &77.7 &48.4 &61.2 \\
        \rowcolor{mygray} MkNet-SV+Ours & Online &\textbf{72.7} &\textbf{84.1} &\textbf{56.7} &\textbf{70.1} \\
        \bottomrule
    \end{tabular}
\end{table}

\begin{table}[t]
    \setlength{\tabcolsep}{4pt}
    \centering
    \small
    \caption{3D object detection and instance segmentation results on ScanNet dataset. Offline and online methods are separated by horizontal line. $^\dag$ means INS-Conv requires an additional 3D reconstruction algorithm to acquire high-quality point clouds or meshes.}\label{tab2}
    \begin{tabular}{c|cc|c|cc}  
        \toprule
        \multicolumn{3}{c|}{Detection} & \multicolumn{3}{c}{Insseg} \\
        \midrule
        \multirow{2}{*}{Method} & \multicolumn{2}{c|}{mAP} & \multirow{2}{*}{Method} & \multicolumn{2}{c}{mAP} \\
        & @25 & @50 & & @25 & @50 \\
        \midrule
        FCAF3D~\cite{rukhovich2022fcaf3d} &70.7 &56.0 &SoftGroup~\cite{vu2022softgroup} &78.9 &67.6 \\
        CAGroup3D~\cite{wang2022cagroup3d} &74.5 &60.3 &TD3D~\cite{kolodiazhnyi2023top} &81.3 &71.1 \\
        \midrule
        AnyView~\cite{wu2023anyview} &60.4 &36.0 &INS-Conv$^\dag$~\cite{liu2022ins} &-- &57.4 \\
        \rowcolor{mygray} FCAF3D-SV &41.9 &20.6 &TD3D-SV & 53.7 &36.8 \\
        \rowcolor{mygray} FCAF3D-SV+Ours &\textbf{70.5} &\textbf{49.9} &TD3D-SV+Ours & \textbf{71.3} & \textbf{60.5} \\
        \bottomrule
    \end{tabular}
\end{table}

\begin{table*}[]
	\centering
	\setlength{\abovedisplayskip}{0pt}
	\setlength{\belowdisplayskip}{0pt}
    \small
	\caption{The performance of different 3D scene perception methods on ScanNet online benchmark. We report mIoU / mAcc, mAP@25 / mAP@50 and mAP@25 / mAP@50 for semantic segmentation, object detection and instance segmentation respectively.}\label{tab3}
	\begin{tabular}{c|l|p{1.5cm}<{\centering}|p{1.5cm}<{\centering}|p{1.5cm}<{\centering}|p{1.5cm}<{\centering}|p{1.5cm}<{\centering}|p{1.5cm}<{\centering}|p{1.5cm}<{\centering}}
		\toprule
		 & \multirow{2}{*}{Method} & \multirow{2}{*}{Type} & \multicolumn{3}{c|}{Number of Sequence} & \multicolumn{3}{c}{Length of Sequence} \\
         & & &\ 1\ &5\ &10 &5\ &10 &15 \\
		\midrule
        \multirow{4}{*}{\rotatebox[origin=c]{90}{Semseg}} &MkNet & Offline &63.7 / 73.5 &62.7 / 72.8 &58.9 / 69.4 &59.3 / 69.8 &63.0 / 73.0 &63.5 / 73.7 \\
        &Fs-A & Online &62.0 / 72.8 &60.6 / 71.7 &60.0 / 71.3 &60.1 / 71.3 &60.7 / 71.8 &61.0 / 72.0  \\
        &MkNet-SV & Online &63.3 / 74.3 &63.3 / 74.3 &63.3 / 74.3 &63.3 / 74.3 &63.3 / 74.3 &63.3 / 74.3  \\
        &MkNet-SV+Ours & Online &\textbf{69.1} / \textbf{82.2} &\textbf{66.8} / \textbf{80.0} &\textbf{65.9} / \textbf{79.2} &\textbf{65.9} / \textbf{79.3} &\textbf{66.8} / \textbf{80.1} &\textbf{67.1} / \textbf{80.4}  \\
		\midrule
        \multirow{4}{*}{\rotatebox[origin=c]{90}{Detection}} &FCAF3D & Offline &57.0 / 40.6 &41.1 / 25.2 &34.6 / 19.3 &28.4 / 15.2 &33.9 / 19.4 &37.7 / 22.8 \\
        &AnyView & Online &60.4 / 36.0 &48.8 / 25.3 &43.1 / 20.5 &36.6 / 16.5 &42.0 / 20.7 &45.6 / 23.8  \\
        &FCAF3D-SV & Online &41.9 / 20.6 &29.8 / 13.3 &27.0 / 11.5 &24.4 / 10.1 &26.2 / 11.0 &27.6 / 12.1 \\
        &FCAF3D-SV+Ours & Online &\textbf{70.5} / \textbf{49.9} &\textbf{58.7} / \textbf{37.7} &\textbf{56.2} / \textbf{34.3} &\textbf{53.1} / \textbf{31.2} &\textbf{54.9} / \textbf{33.8} &\textbf{56.1} / \textbf{35.6}  \\
        \midrule
        \multirow{3}{*}{\rotatebox[origin=c]{90}{Insseg}} &TD3D & Offline & 64.0 / 50.8 &61.6 / 49.7 &59.4 / 48.4 &59.0 / 47.9 &61.4 / 49.8 &61.7 / 49.8 \\
        &TD3D-SV & Online &53.7 / 36.8 &54.2 / 41.6 &57.0 / 46.3 &56.4 / 45.5 &53.9 / 40.9 &52.6 / 39.5 \\
        &TD3D-SV+Ours & Online &\textbf{71.3} / \textbf{60.5} &\textbf{64.7} / \textbf{55.2} &\textbf{64.2} / \textbf{55.0} &\textbf{64.0} / \textbf{54.7} &\textbf{64.6} / \textbf{55.1} &\textbf{63.9} / \textbf{54.3} \\
		\bottomrule
	   \end{tabular}
\end{table*}

\textbf{Room-level benchmarks:} By default, offline methods take in reconstructed point clouds and online methods take in posed RGB-D videos without 3D reconstruction. Special case is denoted by $^\dag$.
Note that there is a challenge in online methods when compare to offline alternatives, as offline methods directly process the complete and clean 3D geometry of rooms while online methods deal with partial and noisy frames. According to Table \ref{tab1} and Table \ref{tab2}, by simply inserting the memory-based adapters into $\mathcal{M}_{SV}$, we significantly boost their accuracy on complete scenes and achieve better performance compared with state-of-the-art online 3D scene perception models specially designed for each task. We observe the improvement upon $\mathcal{M}_{SV}$ is especially significant on 3D object detection and instance segmentation tasks. This is because these tasks require complete predictions for each object, while it is usually very hard to infer the whole geometry of large objects with a single RGB-D frame.
We also notice our method even outperforms offline methods on semantic segmentation task. Since this task requires more detailed perception of local geometry rather than the global context, our method can predict finer segmentation with only partial and noisy inputs.

\textbf{Online benchmark:} In this benchmark, the inputs to all methods are the posed RGB-D sequences. We concatenate the point clouds from each RGB-D frame of a sequence into a whole for offline methods. As the code of INS-Conv is not accessible, we do not compare with it on this benchmark. According to Table \ref{tab3}, offline methods show bad generalization ability on partial and noisy scenes, especially when the input sequence is short. Note that offline methods take in the whole observed scene $S_t$ at each time. When processing $S_{t+1}$, the features extracted for $S_t$ is wasted. On the contrary, online methods process a single frame $x_t$ at each time and fuse the per-frame predictions, which is much more efficient and practical in real-time robotic tasks. Equipped with our memory-based adapters, $\mathcal{M}_{SV}$ achieves the best performance compared with other offline and online methods on all tasks and experimental settings. We observe the longer the input sequence, the larger the improvement upon $\mathcal{M}_{SV}$, which validates our modules can effectively aggregate long-term temporal information.

\begin{table}[]
    \centering
    \setlength\tabcolsep{14pt}
    \caption{Ablation study on point cloud and image modules. We report semantic segmentation results on ScanNet. The performance of image module is based on point cloud module.}\label{tab45}
    \footnotesize
    \begin{tabular}{l|p{0.5cm}<{\centering}p{0.5cm}<{\centering}}
        \toprule
        Method & mIoU & mAcc \\
        \midrule
        Remove residual connection &64.6 &77.9 \\
        Random initialization &66.2 &78.6 \\
        Remove voxel maxpooling &64.8 &76.1 \\
        Set scaling factor $s=1$ &65.3 &78.4 \\
        Set scaling factor $s=5$ &66.8 &79.3 \\
        Insert after neck &66.0 &78.8 \\
        \textbf{The final point cloud module} &\textbf{66.9} &\textbf{79.3} \\
        \midrule
        Remove residual connection &67.1 &79.8 \\
        Random initialization &68.7 & 81.7 \\
        Set shift ratio $\tau=4$ &68.9 &82.1 \\
        Set shift ratio $\tau=16$ &68.7 &81.9 \\
        Remove 3D to 2D adapter &68.0 &80.8 \\
        Insert after neck &68.4 &81.6 \\
        \textbf{The final image module} &\textbf{69.1} &\textbf{82.2} \\
        \bottomrule
    \end{tabular}
\end{table}

\begin{table}[]
    \centering
    \setlength\tabcolsep{8pt}
    \caption{Effects of our memory-based adapters when both image and point cloud backbones are fixed during finetuning.}\label{tab6}
    \footnotesize
    \begin{tabular}{c|c|c|c}
        \toprule
         & MkNet & FCAF3D & TD3D \\
        \midrule
        Fix I &69.1 / 82.2 &70.5 / 49.9 &71.3 / 60.5 \\
        Fix P \& I &67.3 / 79.9 &66.4 / 47.1 &69.1 / 58.2 \\
        \bottomrule
    \end{tabular}
\end{table}

\begin{figure*}[t]
    \centering
    \vspace{-1mm}
    \includegraphics[width=\linewidth]{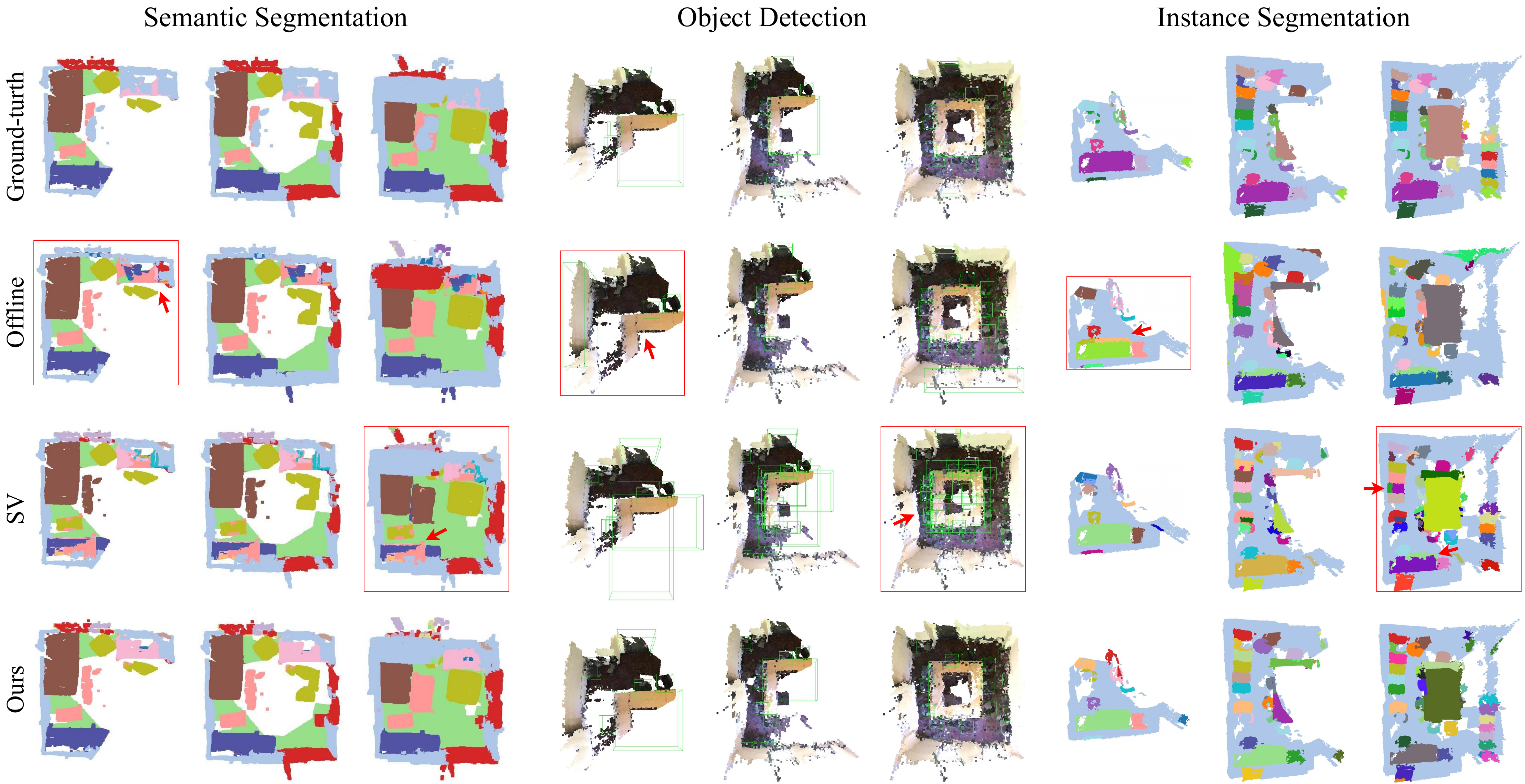}
    \caption{Visualization results on the online benchmark. Our predictions are accurate and robust to the number of frames. Note that some ground-truth masks are incomplete due to the noisy 2D annotations, in this case our predictions are more reasonable than the ground-truths.}
    \label{vis}
\end{figure*}

We visualize the predictions of different methods in Figure \ref{vis}. 
It can be seen that our method is more accurate than $\mathcal{M}_{SV}$ due to the temporal modeling ability, and more robust to number of frames than offline methods.

\subsection{Ablation Study}
We first ablate the design choices of two memory-based adapters on 3D semantic segmentation task on ScanNet. 
Besides, we further show the performance of our method when both image and point cloud backbones are fixed during finetuning the adapters. 

\textbf{Point cloud and image modules:} Table \ref{tab45} validates the effectiveness of our designs. We observe removing voxel maxpooling significantly degrades the performance, which shows the importance of updating memory. With the increase of $s$, the performance first improves and then keeps steady or even slightly declines, which indicates the neighbor context information is important for temporal learning, but too large neighbor voxel set brings much redundant features. 
Large $s$ will also increase the computation overhead, so we choose $s=2.5$ to achieve the best accuracy-computation tradeoff.
We observe the influence of $\tau$ is similar with $s$ and thus choosing a proper value is important for both high accuracy and less memory storage. From these experiments, we also validate the effectiveness of the 'adapter paradigm', which includes residual connection, zero-initialization and inserting after backbone.

\textbf{Fixed backbones:} When finetuning our adapters, we fix the image backbone and finetune other parameters. We further study the effects of our method when both image and point cloud backbones are fixed. As shown in Table \ref{tab6}, even with both image and point cloud backbones fixed, our method still achieves state-of-the-art performance on all three online tasks. In this way, we can further reduce the memory footprint and training time, which provides the users with more efficiency-accuracy tradeoff.
  
\section{Conclusion}
  In this paper, we have presented memory-based adapters for online 3D scene perception. Mainstream 3D scene perception methods are offline, which is hard to be applied in most real-time applications where only streaming RGB-D video is accessible. Existing online perception methods design model and task-specific temporal learning approaches, but most of them only focus on temporal aggregation for single modality and thus cannot fully exploit temporal relations between image and point cloud features. 
To this end, we propose plug-and-play temporal learning modules, which can empower offline methods with online perception ability by simply inserting memory-based adapters and finetuning on RGB-D videos. Specifically, given point cloud and image features extracted from the backbones, we first devise a queued memory mechanism to cache these information over time and maintaining a reasonable storage overhead. Then we devise aggregation modules which directly operate on the memory and pass temporal information from the cached features to current frame. As the global context of image features is limited due to the short queue, we further propose 3D-to-2D adapter to enhance image features with 3D memory.
We conduct extensive experiments on ScanNet and SceneNN. By equipping offline models with our modules, we achieve leading performance on three scene perception tasks compared with state-of-the-art online methods, even without any model and task-specific designs.

\clearpage
\appendix
\section*{Supplementary Material}
\noindent This supplementary material is organized as follows: 
\begin{itemize}
    \item Section \ref{architecture} demonstrates the detailed architecture of our baseline models in three tasks and how to insert our adapters into them.
    \item Section \ref{hyperparameters} details the training hyperparameters adopted in our experiments. 
    \item Section \ref{category} details per-class experimental results. 
\end{itemize}

\section{Detailed Architecture}\label{architecture}
\begin{figure}[t]
    \centering
    \includegraphics[width=1.0\linewidth]{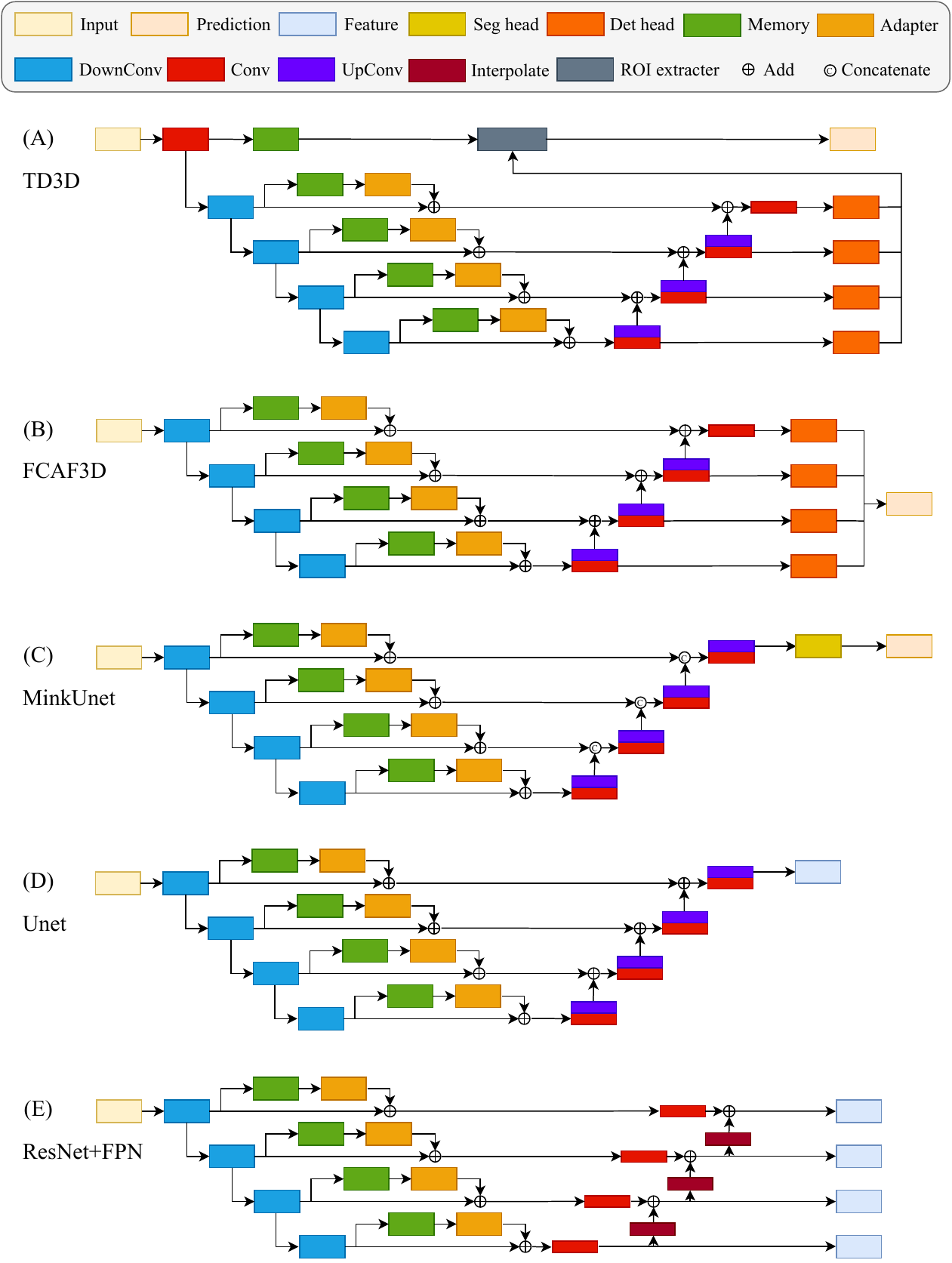}
    \caption{Details about the architectures of image and point cloud backbones and how to insert the adapters into them.}
    \label{arch}
\end{figure}

We illustrate the architectures of both image and point cloud backbones and show how to insert the memory-based adapters into them in Figure \ref{arch}.
For online 3D semantic segmentation, we use U-Net~\cite{ronneberger2015u} as the image backbone and Minkowski-UNet~\cite{choy20194d} as the point cloud backbone, which is shown in Figure \ref{arch} (D) and (C) respectively.
For online 3D object detection, we adopt ResNet~\cite{he2016deep} with FPN~\cite{lin2017fpn} as the image backbone and FCAF3D~\cite{rukhovich2022fcaf3d} as the point cloud backbone, which is shown in Figure \ref{arch} (E) and (B) respectively.
For online 3D instance segmentation, we use the same image backbone as the object detection task and adopt TD3D~\cite{kolodiazhnyi2023top} as the point cloud backbone, which is shown in Figure \ref{arch} (E) and (A) respectively.
Note that for TD3D, the backbone maintains a high-resolution scene representation for ROI-wise instance prediction. We consrtuct a point cloud memory to cache this scene representation, which ensures the point clouds within each ROI are the most complete up to current time. This design helps us acquire complete instance mask by simply performing 3D NMS, which avoids complicated mask fusion strategy~\cite{liu2022ins} to merge instance masks of different frames.

\section{Training Hyperparameters}\label{hyperparameters}
We train the online perception models in two stage. Firstly we train single-view perception model $\mathcal{M}_{SV}$ on ScanNet-25k~\cite{dai2017scannet}. 
Secondly we insert the memory-based adapters into $\mathcal{M}_{SV}$ and finetune the network on ScanNet RGB-D videos.

For online semantic segmentation, we set max epoch as 250, weight decay as 0.01, initial learning rate as 0.0008 and adopt AdamW optimizer with OneCycleLR scheduler for the first stage. Then we set max epoch as 36, weight decay as 0.01, initial learning rate as 0.008 and adopt AdamW optimizer with a stepwise scheduler which steps at 24 and 32 epoch for the second stage.

For online object detection, we set max epoch as 12, weight decay as 0.0001, initial learning rate as 0.001 and adopt AdamW optimizer with a stepwise scheduler which steps at 8 and 11 epoch for the first stage. Then we adopt the same hyperparameters for finetuning in the second stage.

For online instance segmentation, we set max epoch as 33, weight decay as 0.0001, initial learning rate as 0.001 and adopt AdamW optimizer with a stepwise scheduler which steps at 28 and 32 epoch for the first stage. Then we adopt the same hyperparameters for finetuning.

\section{Class-specific Results}\label{category}
We provide class-specific experimental results of out method on three 3D scene perception tasks. Table \ref{supp:tab1} and \ref{supp:tab2} show the 3D semantic segmentation results on ScanNet and SceneNN dataset with per-class IoU. 
Table \ref{supp:tab3} and \ref{supp:tab4} show the 3D object detection results on ScanNet dataset with per-class AP$_{25}$ and AP$_{50}$.
Table \ref{supp:tab5} and \ref{supp:tab6} show the 3D object detection results on ScanNet dataset with per-class AP$_{25}$ and AP$_{50}$.

\newpage

\begin{table*}[tp]
	\centering
	\caption{Per-class 3D semantic segmentation results (IoU) of our method on the ScanNet validation set.}
        \vspace{-6pt}
	\scalebox{0.8}{\setlength{\tabcolsep}{0.3mm}{
	\begin{tabular}{@{}cccccccccccccccccccccc@{}}
		\toprule
		     & wall   & floor & cabinet & bed   & chair & sofa  & table & door  & window & bookshelf & picture &         counter & desk  &  curtain & fridge & curtain & toilet & sink  & bathtub & others & mean \\ 
            \midrule
            Ours   & 85.7  & 97.1 & 63.1  & 80.7 & 89.0 & 76.1 & 73.7   & 66.1  & 63.6    & 77.1     & 41.8     & 65.5   & 61.2 & 58.9   & 61.0       & 72.7    & 95.2  & 77.9 & 94.2   & 53.6 &72.7    \\     
            \bottomrule

    \end{tabular}}}
	\label{supp:tab1}\vspace{-8pt}
\end{table*} 

\begin{table*}[tp]
	\centering
	\caption{Per-class 3D semantic segmentation results (IoU) of our method on the SceneNN validation set.}
        \vspace{-6pt}
	\scalebox{0.8}{\setlength{\tabcolsep}{0.3mm}{
	\begin{tabular}{@{}cccccccccccccccccc@{}}

		\toprule
		     & wall   & floor & cabinet & bed   & chair & sofa  & table & door  & window & bookshelf & picture &         counter & desk  &  curtain & fridge  & sink  & mean \\ 
            \midrule
            Ours   & 75.3  & 82.6 & 59.8  & 82.8 & 62.0 & 57.8 & 18.4   & 52.4  & 20.5    & 55.9     & 29.4     & 52.6   & 44.9 & 50.2   & 80.3       & 81.8    & 56.7   \\     
            \bottomrule
    \end{tabular}}}
	\label{supp:tab2}\vspace{-8pt}
\end{table*}

\begin{table*}[tp]
	\centering
	\caption{Per-class 3D object detection results (AP$_{25}$) of our method on the ScanNet validation set.}
        \vspace{-6pt}
	\scalebox{0.8}{\setlength{\tabcolsep}{0.3mm}{
	\begin{tabular}{@{}ccccccccccccccccccccc@{}}
		\toprule
		  & cabinet & bed   & chair & sofa  & table & door  & window & bookshelf & picture & counter & desk  &  curtain & fridge & curtain & toilet & sink  & bathtub & others&mean \\ \midrule
        Ours           & 55.2  & 85.4 & 88.7  & 87.2 & 63.3 & 62.5 & 47.3   & 66.2  & 36.0    & 65.2     & 80.1     & 65.0   & 58.1 & 76.3   & 99.7       & 76.7    & 93.3  & 62.0 &70.5   \\     \bottomrule
    \end{tabular}}}
	\label{supp:tab3}\vspace{-8pt}
\end{table*}

\begin{table*}[tp]
	\centering
	\caption{Per-class 3D object detection results (AP$_{50}$) of our method on the ScanNet validation set.}
        \vspace{-6pt}
	\scalebox{0.8}{\setlength{\tabcolsep}{0.3mm}{
	\begin{tabular}{@{}ccccccccccccccccccccc@{}}
		\toprule
		  & cabinet & bed   & chair & sofa  & table & door  & window & bookshelf & picture & counter & desk  &  curtain & fridge & curtain & toilet & sink  & bathtub & others&mean \\ \midrule
        Ours           & 36.7  & 75.6 & 73.9  & 77.9 & 57.0 & 33.8 & 19.8   & 43.7  & 19.4    & 26.3     & 62.8     & 32.4   & 41.1 & 24.6   & 89.2       & 46.7    & 84.8  & 52.2 &49.9   \\     \bottomrule
    \end{tabular}}}
	\label{supp:tab4}\vspace{-8pt}
\end{table*}

\begin{table*}[tp]
	\centering
	\caption{Per-class 3D instance segmentation results (AP$_{25}$) of our method on the ScanNet validation set.}
        \vspace{-6pt}
	\scalebox{0.8}{\setlength{\tabcolsep}{0.3mm}{
	\begin{tabular}{@{}ccccccccccccccccccccc@{}}
		\toprule
		  & cabinet & bed   & chair & sofa  & table & door  & window & bookshelf & picture & counter & desk  &  curtain & fridge & curtain & toilet & sink  & bathtub & others&mean \\ \midrule
        Ours           & 60.3  & 86.8 & 91.5  & 80.3 & 72.8 & 56.0 & 55.3   & 67.5  & 45.1    & 48.9     & 72.9     & 68.4   & 56.5 & 86.3   & 99.7       & 81.3    & 87.8  & 65.3 &71.3   \\     \bottomrule
    \end{tabular}}}
	\label{supp:tab5}\vspace{-8pt}
\end{table*}

\begin{table*}[tp]
	\centering
	\caption{Per-class 3D instance segmentation results (AP$_{50}$) of our method on the ScanNet validation set.}
        \vspace{-6pt}
	\scalebox{0.8}{\setlength{\tabcolsep}{0.3mm}{
	\begin{tabular}{@{}ccccccccccccccccccccc@{}}
		\toprule
		  & cabinet & bed   & chair & sofa  & table & door  & window & bookshelf & picture & counter & desk  &  curtain & fridge & curtain & toilet & sink  & bathtub & others&mean \\ \midrule
        Ours           & 50.9  & 79.1 & 82.5  & 71.3 & 63.6 & 44.0 & 36.0   & 45.5  & 38.5    & 30.3     & 57.3     & 49.8   & 52.9 & 78.9   & 99.7       & 66.6    & 84.9  & 56.9 & 60.5   \\     \bottomrule
    \end{tabular}}}
	\label{supp:tab6}\vspace{-8pt}
\end{table*}

{
    \small
    \bibliographystyle{ieeenat_fullname}
    \bibliography{main}
}

\end{document}